%% file: Formatting-Instructions-LaTeX-2022.tex
\def\year{2022}\relax
\documentclass[letterpaper]{article} 
\usepackage{aaai22}  
\usepackage{times}  
\usepackage{helvet}  
\usepackage{courier}  
\usepackage[hyphens]{url}  
\usepackage{graphicx} 
\usepackage{natbib}  
\usepackage{caption} 
\DeclareCaptionStyle{ruled}{labelfont=normalfont,labelsep=colon,strut=off} 
\usepackage{algorithm}
\usepackage{algorithmicx}
\usepackage[noend]{algpseudocode}

\usepackage{soul}
\usepackage[hidelinks]{hyperref}
\usepackage{amsmath}
\usepackage{booktabs}
\usepackage{bm}
\usepackage{amssymb}
\usepackage{subfigure} 

\usepackage{newfloat}
\usepackage{listings}
\floatstyle{ruled}
\newfloat{listing}{tb}{lst}{}
\floatname{listing}{Listing}
\title{Active Learning for Event Extraction with Memory-based Loss Prediction Model}
\author{
Shirong Shen ,
Zhen Li ,
Guilin Qi\\
School of Computer Science and Engineering, Southeast University, China\\
\{ssr, jeffery.lee.0628, gqi\}@seu.edu.cn

}

\affiliations{

}

\usepackage{bibentry}

\begin{document}

\maketitle

\begin{abstract}
Event extraction (EE) plays an important role in many industrial application scenarios, and high-quality EE methods require a large amount of manual annotation data to train supervised learning models.
However, the cost of obtaining annotation data is very high, especially for annotation of domain events, which requires the participation of experts from corresponding domain.
So we introduce active learning (AL) technology to reduce the cost of event annotation.
But the existing AL methods have two main problems, which make them not well used for event extraction.
Firstly, the existing pool-based selection strategies have limitations in terms of computational cost and sample validity.
Secondly, the existing evaluation of sample importance lacks the use of local sample information.
In this paper, we present a novel deep AL method for EE.
We propose a batch-based selection strategy and a Memory-Based Loss Prediction model (MBLP) to select unlabeled samples efficiently.
During the selection process, we use an internal-external sample loss ranking method to evaluate the sample importance by using local information.
Finally, we propose a delayed training strategy to train the MBLP model.
Extensive experiments are performed on three domain datasets, and  our method outperforms other state-of-the-art methods.

\end{abstract}

\input{introduction}

\input{related_work}
\input{define}
\input{method.tex}
\input{experiment}

\input{conclusion}

\bibliography{aaai22}
\bibliographystyle{aaai22}

\end{document}

%% file: introduction.tex
\section{Introduction}
Event extraction is an important task in information extraction, including identification and classification of trigger words, argument identification and argument role classification.\cite{chen2015event}.
Trigger word is the main word that most clearly expresses the occurrence of an event.
Argument is an entity temporal expression or value that is involved in an event.
Argument role is the relationship between an argument to the event in which it participates.
For example, in the following sentence:
\textit{In Baghdad, a \underline{cameraman} \textbf{died} when an American tank \textbf{fired} on the Palestine hotel.}
\textit{\textbf{died}} and \textit{\textbf{fired}} are the event triggers for the events of types \textit{Die} and \textit{Attack}, respectively.
\textit{\underline{cameraman}} is an argument in the \textit{Die} event, and the role of \textit{\underline{cameraman}} is \textit{Victim}.

Current state-of-the-art methods are based on supervised learning\cite{yubo2015event,sha2018jointly,yang2019exploring}, which requires a large number of manual labeling samples as supervision data\cite{ren2020survey}.
However, the cost of manually labeling supervision data of EE is very high, especially for event extraction in specific domains\cite{nuij2013automated,capet2008risk,abrahams2002developing,vanegas2015overview,shen2020hierarchical}, such as financial, legal, military.
Therefore, it is necessary to reduce the amount of supervision data needed for training event extraction models.

Active learning (AL)\cite{culotta2005reducing} aims at reducing the amount of supervision data annotated by the human expert by designing a sample selection strategy\cite{settles2009active,cohn1994improving}.
After evaluating the importance of samples in the unlabeled sample pool, AL selects a few samples for supervised learning.
Considering the overconfidence of the deep neural networks  \cite{wang2014new} and the uncertainty of feature coding\cite{wang2016cost}, an end-to-end loss prediction model is used to evaluate the importance of unlabeled samples\cite{yoo2019learning}.
Currently, there are few researches applying AL technology to EE in specific domain with heuristic selection strategies\cite{han2016active,lybarger2020annotating,wonsild2020danish}.
These works show the effectiveness of AL in reducing the costs of labeling data.
%

However, as far as we know, there are two problems when existing AL methods are applied to EE:
(1) The existing pool-based selection strategies have limitations  in calculating cost and sample validity. 
Pool-based selection strategy includes optimal set selection and greedy selection.
The computational cost of optimal set selection is extremely high, so it is difficult to apply it to the deep neural network model.
Greedy selection may result in the excessive selection of samples of the same category, resulting in redundant labeling\cite{yoo2019learning}.
(2) Existing methods for evaluating the importance of samples lack the use of local information of samples.
Each sample in EE has multiple predictions for multiple tasks. 
This requires the sample selection strategy to integrate all the predictions to evaluate the importance of the sample.
However, the existing methods ignore the influence of local information on sample importance.

To solve the problem (1), we propose a new batch-based selection strategy and a novel Memory-Based Loss Prediction (MBLP) model for AL based EE.
Batch-based selection strategy divides the unlabeled sample pool into several batches, and only the most important samples are selected from each batch according to the loss prediction results.
When making loss prediction, MBLP will use the information of the sample selected in previous batches.
To solve the problem (2), we design an inter-exter sample loss ranking method to compare the importance of different samples.
For an unlabeled sample, we select several predictions with the greatest loss as the evaluation basis of its importance, and then compare it with other samples.
This method can balance the performance of sub-tasks while paying attention to the local information of the sample.  
In addition, we also propose a new delayed training strategy, which trains MBLP model and EE models simultaneously by supervised learning.
The main contributions of this paper are summarized as follows:
\begin{itemize}
    \item We propose a new batch-based selection strategy and a new memory-based loss prediction (MBLP) model, which breaks through the limitations of existing sample selection methods in terms of computational cost and global quality. We design a delayed training strategy to train MBLP, which simulates the sample selection process of AL as the training process.
    \item We propose an inter-exter sample loss ranking method to measure the importance of samples. This method makes up for the problem that previous methods ignored the local information of samples in the importance evaluation stage, and is more suitable for AL based EE task.
    \item We evaluate our model on event extraction datasets from three domains. The experimental results show that our method can not only ensure the effectiveness of EE model, but also reduce the amount of supervision data needed for training more effectively than other AL methods.
\end{itemize}


%% file: related_work.tex
 \section{Related Work}

\subsection{Event Extraction}

Event extraction(EE) is a crucial information extraction(IE) task that aims to extract event information in texts, which has attracted extensive attention among researchers. Traditional EE methods employ manually-designed features, such as the syntactic feature\cite{ahn2006stages} document-level feature\cite{ji2008refining} and other features \cite{liao2010using,li2013joint} for the task. Modern EE methods employ neural models, such as Convolutional Neural Networks\cite{chen2015event},  Recurrent Neural Networks\cite{nguyen2016joint,sha2018jointly}, Graph Convolutional Neural Networks\cite{liu2018jointly,liu2019neural} and other advanced architectures\cite{yang2016joint,liu2018event,liu2019exploiting,nguyen2019one,liu2020event}.


EE is also an important research direction in specific domains. In financial domain, EE can help companies quickly discover product markets, analyze risk and recommend transactions\cite{nuij2013automated,capet2008risk,abrahams2002developing}. In biomedical domain, EE is used to identify the change of biomolecule state and the interaction between biomolecules, so as to understand physiology and pathogenesis\cite{vanegas2015overview}. Shen et al.\cite{shen2020hierarchical} use EE technology to quickly and accurately capture the key information in legal cases, providing technical support for accurate case retrieval and recommendation.

At present, EE methods need a lot of labeled data for supervised learning, and the labeling cost becomes an important obstacle for these methods to put into practical production.

\subsection{Active Learning}
In order to reduce the need for labeled samples in supervised learning, active learning (AL) uses a well-designed sample selection strategy to select few sample from the unlabeled sample pool for labeling.
Common sample selection strategies include uncertainty-based methods\cite{beluch2018power,joshi2009multi,lewis1994sequential,ranganathan2017deep,tong2001support}, diversity-based methods\cite{bilgic2009link,gal2017deep}, expected model changes\cite{freytag2014selecting,chitta2019training}, and committee query methods\cite{seung1992query}. 
In addition, there are some methods to select samples for labeling by taking uncertainty and sample diversity as standards\cite{shui2020deep,yin2017deep,zhdanov2019diverse,manning2015computational}.

To adapt to deep neural networks, deep AL was first applied to image classification\cite{gal2017deep,kendall2017uncertainties}. 
Recently, some researches have studied the deep AL methods in natural language processing(NLP) and IE, including sentence classification \cite{zhang2017active}, named entity recognition\cite{shen2017deep,chen2015study,gao2019recognizing}, and event extraction\cite{han2016active,lybarger2020annotating,wonsild2020danish}. 

However, the current AL method has some limitations in calculating cost and evaluating sample importance, so it can not be well applied to the EE task.


%% file: define.tex
\section{Preliminaries}
\subsection{Event Extraction}
Given a sentence $S = \{w_1,w_2,\dots,w_n \}$, where $n$ is the sentence length and $w_i$ is the $i$-th token of $S$. The event extraction model detects all event triggers with specified event types in $S$, and extracts all arguments corresponding to each trigger with argument roles. 
We denote $Y=\{y\}_M$ as the set of target event types, where $M$ is the number of event types. 
In order to reduce unnecessary labeling costs and reduce computational complexity, we enumerate every \emph{noun}, \emph{verb}, and \emph{adjective} in the sentence as candidate triggers. 
All these words can be identified by natural language analysis tools.
Then we consider trigger extraction as a word-level classification task.
Let $T_S = \{tr_1, tr_2, \dots , tr_k \}$ be the set of candidate trigger words of $S$, the event extraction model predicts the event type corresponding to each candidate trigger word in $Y$.
\textit{NA}$\in Y$ is a specific type means the word is not an event trigger in $S$.
We denote the set of target argument roles as $R=\{r\}_N$, where $N$ is the number of argument roles.
We consider argument extraction for each trigger as sequence label task, and apply BIO annotation schema to assign a label to each token $w_i$. 
Each $r \in R$ corresponds to two labels($B$-$r$ and $I$-$r$) to indicate the token span corresponding to this argument role, where $B$-$r$ means the starting token of an argument with role $r$, $I$-$r$ represent the inside token of an argument with role $r$.
$O$ means the token does not belong to any argument.
All labels of argument extraction form set $R^{ar}=\{r^{ar}\}_{2 \times N + 1}$. 

\subsection{Active Learning}
\label{sec:active}
In AL, all data is divided into two sets.
Let $\mathcal{L}=\{(S,T_S,L_S)\}$ be the dataset of labeled sentence, $\mathcal{U}=\{(S',T_{S'})\}$ denotes the unlabeled data pool.

$L_S = (L_{ST},L_{SA})$ consists of two parts. $L_{ST} = \{y_1, \dots , y_k\}$ is the label set of candidate trigger $T_S$, where $y_i \in Y$. 
$L_{SA} = \{r^{ar}_{i,j},1\leq i \leq k,1 \leq j \leq n \}$ is the label set of argument extraction, where $r^{ar}_{i,j}$ means the label of the $w_j$ when the $tr_i$ is the trigger word, if the label of $tr_i$ is \textit{NA}, then $\forall j, r^{ar}_{i,j} = O$ . 

In the AL process, we will select samples from $\mathcal{U}$ multiple rounds for labeling and add them to $\mathcal{L}$.
We denote the selected sample set for $t$-th round as $U_t$, $\mathcal{U}=\mathcal{U} \backslash U_t$, which means to remove the samples in $U_t$ from $\mathcal{U}$.
$L_t = label(U_t)$ denotes that $L_t$ is the labeled sample set of $U_t$.
$\mathcal{L} = \mathcal{L} \cup L_t$ refers to adding the labeled samples in $L_t$ to set $\mathcal{L}$.
We first randomly select $U_0$ from $\mathcal{U}$ to label it as $\mathcal{L}$. Subsequently, $\mathcal{L}$ is used for supervised learning, and each time the model converges, the next sample selection and labeling are performed until the maximum number of selections is reached. 

%% file: method.tex
\section{Methodology}
\begin{figure}[hbp]
    \centering
    \includegraphics[width=0.4\textwidth]{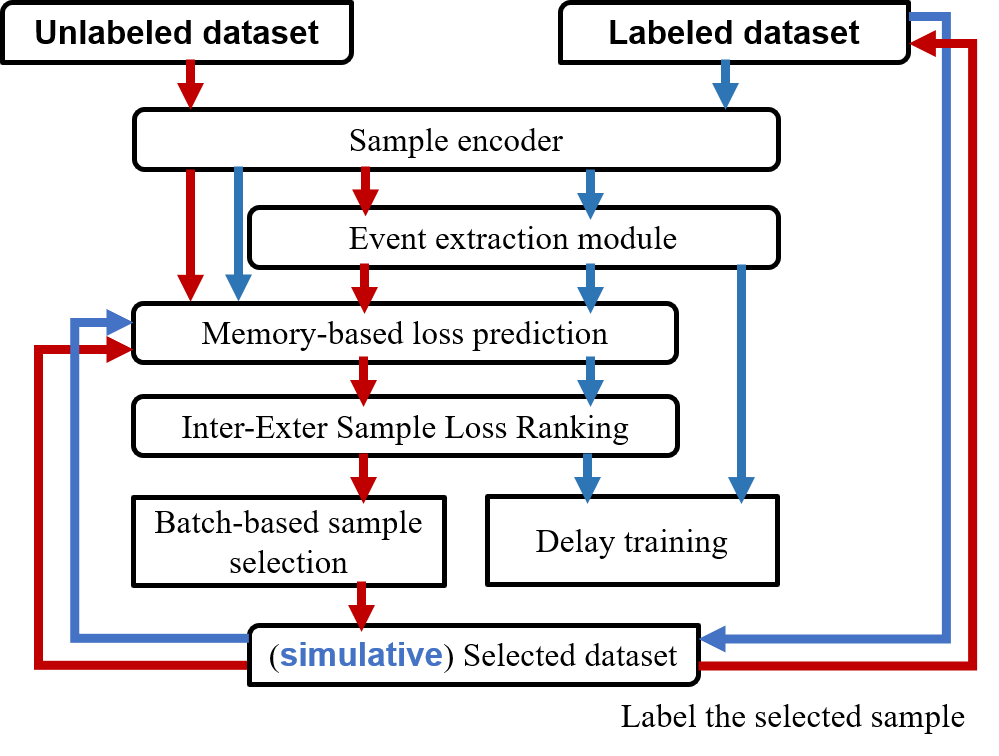}
    \caption{The overall framework of our active learning based event extraction method. The red line represents the sample selection process, and the blue line represents the supervised learning process. The training process is shown in Figure \ref{fig:delayed}. }
    \label{fig:frame}
\end{figure}
This section introduces our AL based EE framework.As shown in Figure \ref{fig:frame}, the entire framework consists of the following modules: 
(1) \textbf{Sample encoder}: a Bert-based encoder is used to encode sample into real-valued vectors, which are shared by other modules. 
(2) \textbf{Event extraction module}: event extraction is split into two classification tasks, namely trigger classification and argument role classification. 
(3) \textbf{Memory-based loss prediction module}: a memory-based networks is used to predict the loss of each predictions in samples. 
(4) \textbf{Sample selection strategy}: we design a batch-based sample selection strategy and evaluate the importance of unlabeled samples by an inter-exter sample loss ranking method. 
(5) \textbf{Delayed training process}: we customize a new delayed training method. With the delayed training method, loss prediction module and event extraction model can be trained together and predict independently.

\subsection{Sample Encoder}
Given a sentence $S$, we use a pre-trained Bidirectional Encoder Representations from Transformers (BERT) \cite{devlin2018bert} to convert the sentence into a vector sequence.
First, each token in $S$ is transformed into a fixed-length token embedding, and its position embedding depend on its position in $S$.
The input of BERT is the concatenation of token embedding and position embedding.
After the encoding of a pre-trained bidirectional transformer, each token is transformed into a feature vector.
We use $\mathcal{B}(S,w_i)$ to denote the feature vector of $w_i$ in sentence $S$ generated by BERT.

\subsection{Event Extraction Module}
We adopt the pre-trained language model based joint extraction schema as other EE models\cite{yang2019exploring,shen2020hierarchical}.
For trigger extraction, we formulate it as a word-level classification task with labels being event type in $Y$. We use a multi-classifier after BERT as trigger extractor.
For each candidate trigger $tr_i$ in sentence $S$, we use the feature vector of $tr_i$ as the input of trigger extractor.
\begin{equation}
\setlength\abovedisplayskip{0.1cm}
\setlength\belowdisplayskip{0.1cm}
    \mathbf{P}^i_{tr}(S,tr_i) = C_{tr}(\mathcal{B}(S,tr_i))
\end{equation}
where $C_{tr}$ is a neural classifier and $\mathbf{P}^i_{tr}(S,tr_i)$ is a $M$-dimensional vector, each element in $\mathbf{P}^i_{tr}(S,tr_i)$ represents the probability of $tr_i$ belonging to an event type.
In many cases, the trigger is a phrase, so we denote that $\mathcal{B}(S,tr_i)$ is the mean value of the feature vectors of all tokens in $tr_i$.

Since we cannot automatically give the set of candidate arguments, we use sequence labeling to extract arguments.
For each trigger-token pair $(tr_i,w_j)$, we use a multi-classifier as the argument extractor to predict the label of $w_j$ corresponding to trigger $tr_i$. 
The input of the argument extractor includes three parts: the coding vector of the trigger word $tr_i$, the coding vector of the current token $w_j$, and the context feature of $tr_i$ and $w_j$.
We use a multi-head attention mechanism\cite{vaswani2017attention} to generate the context feature of $tr_i$ and $w_j$, where the query in attention is $[\mathcal{B}(S,tr_i),\mathcal{B}(S,w_j)]$, key and value in attention is the encoding of $S$.
\begin{equation}
\setlength\abovedisplayskip{0.1cm}
\setlength\belowdisplayskip{0.1cm}
    c_{i,j} = Att([\mathcal{B}(\!S,tr_i\!)\!,\mathcal{B}(\!S,w_j\!)],\mathcal{B}(\!S\!),\mathcal{B}(\!S\!))
\end{equation}
Where $Att$ represents the multi-head attention mechanism. Formally, the sequence label classifier is as follows,
\begin{equation}
\setlength\abovedisplayskip{0.1cm}
\setlength\belowdisplayskip{0.1cm}
    \mathbf{P}^{i,j}_{ar}(S,r^{ar}_{tr_i,w_j}) = C_{ar}(\mathcal{B}(\!S,tr_i\!),\mathcal{B}(\!S,w_j\!),c_{i,j})
\end{equation}
where $C_{ar}$ is a neural classifier and $\mathbf{P}^{i,j}_{ar}(S,r^{ar}_{tr_i,w_j})$ is a $(2 \times N +1)$-dimensional vector which is the final output of sequence label task, each dimension of $\mathbf{P}^{i,j}_{ar}(S,r^{ar}_{tr_i,w_j})$ corresponds to a BIO label of an argument role.
In the training process of the event extraction model, we use the cross-entropy loss of all tasks as the objective function for training, denoted as $\mathcal{L}_{ee}^S$.

\subsection{Memory-Based Loss Prediction Module}
Existing end-to-end methods can only predict loss of one sample at a time, and can not guarantee the overall quality of a set of samples.
Therefore, we propose a memory-based loss prediction model, which predicts sample loss according to the current state of the model and the information in the currently selected unlabeled samples.

At $t$-th selection round, it is assumed that we have selected some unlabeled samples to join $U_t$, we first store the information in $U_t$ into \emph{selected memory module} (SMM).
Since both trigger word extraction and argument extraction are multi-classification tasks, we will introduce our method on an abstract multi-classification task $\mathcal{T}$.
SMM stores a matrix $\mathbf{M}_{\mathcal{T}}$ for each task $\mathcal{T}$, if the task has $K$ categories, $\mathbf{M}_{\mathcal{T}}$ is a $K \times d_M$ matrix.
The $p$-th row vector of $\mathbf{M}_{\mathcal{T}}$ represents the $p$-th category information from selected unlabeled samples, which is denoted as $\mathbf{M}_{\mathcal{T}}^p$ .

We first calculate the sample's contribution to each category, and then update the memory module with a gate-unit.
Given a unlabeled sample $S$, the update process of SMM is as follows:

\begin{equation}
\label{SMM}
\setlength\abovedisplayskip{0.1cm}
\setlength\belowdisplayskip{0.1cm}
\begin{split}
     \mathbf{M}_{\mathcal{T}}^{p,new} & = 
      g^p_{\mathcal{T}}(W_{\mathcal{T}}^{M} \! f_{\mathcal{T}}^p)   \!+\!(1\!-\!g^p_{\mathcal{T}})\mathbf{M}_{\mathcal{T}}^{p,old} 
       \\
     f_{\mathcal{T}}^p &= Att(\mathbf{M}_{\mathcal{T}}^{p,old} ,\mathcal{B}(S),\mathcal{B}(S)) 
     \\
     g^p_{\mathcal{T}} &=  Sigmoid(W^g_{\mathcal{T}}[f_{\mathcal{T}}^p,\mathbf{M}_{\mathcal{T}}^{p,old}]+b^g_{\mathcal{T}}) 
\end{split}
\end{equation}

where $W_{\mathcal{T}}^{M}$, $W^g_{\mathcal{T}}$, and $b^g_{\mathcal{T}}$ are the learnable neural network parameters, $f_{\mathcal{T}}^p$ is the feature vector of the $p$-th category contained in sample $S$.
The initial state of $\mathbf{M}_{\mathcal{T}}$ is also set as a learnable parameter and updated during the supervision process of the loss prediction model.
In a round of sample selection process, whenever a sample is selected, its information will be stored in SMM. 

After the memory module is updated, we will predict the loss of unlabeled samples. 
Given $S_{\mathcal{T}}^i$, the $i$-th prediction of task $\mathcal{T}$ in sample $S$, we use the following three part information to predict its loss: 
$h^i_{\mathcal{T}}$: the hidden vector of current prediction. For example, if $\mathcal{T}$ is trigger classification, $h^i_{\mathcal{T}}$ is $\mathcal{B}(S,tr_i)$;
$P^i_{\mathcal{T}}$: the output of current model. For example, if $\mathcal{T}$ is trigger classification, $P^i_{\mathcal{T}}$ is $\mathbf{P}^i_{tr}(S,tr_i)$;
$h^i_{\mathbf{M}_{\mathcal{T}}}$: the impact feature of $U_t$ on current prediction, where $h^i_{\mathbf{M}_{\mathcal{T}}}=Att(h^i_{\mathcal{T}},\mathbf{M}_{\mathcal{T}},\mathbf{M}_{\mathcal{T}})$.
The loss prediction of $S_{\mathcal{T}}^i$ is,

\begin{equation}
\setlength\abovedisplayskip{0.1cm}
\setlength\belowdisplayskip{0.1cm}
\begin{split}
     L_{pred}(S_{\mathcal{T}}^i) = F([h^i_{\mathcal{T}};P^i_{\mathcal{T}};h^i_{\mathbf{M}_{\mathcal{T}}}])
\end{split}
\end{equation}

\subsection{Sample Selection Strategy}
We propose a batch-based sample selection strategy to select samples one by one based on loss prediction results.
In the $t$-th round of sample selection, we first initialize $U_t=\emptyset$, and then randomly split $\mathcal{U}$ into a number of fixed-size batches.
Only the most important sample in a batch will be added to $U_t$. 
Whenever a sample is added to $U_t$, it is used to update SMM, so that the information of the previously selected sample can be taken into account in the next sample selection.
Compared with optimal selection and greedy selection, batch-based selection is more efficient and reasonable, which can reduce the comparison times of samples and obtain high-quality $U_t$.

We propose an inter-exter sample loss ranking strategy to select the most important sample in each batch.
Given an unlabeled sample, there are multiple predictions for multiple tasks.
First, we divide all predicted losses by $log(K)$ and get balanced losses, where $K$ is the number of categories corresponding to each prediction.
The purpose of this operation is to balance the results of cross-entropy loss in different classification tasks.
Then we sort the balanced losses of the predictions in one sample, and select the largest $m$ among them as the representative of the current sample.
The mean value of the $m$ losses of predictions as the sample's importance to compare with other samples, and select the sample with the largest loss for labeling.
This method can retain the effective local information in the sample without being affected by noise. 

\subsection{Delayed Training Process}

\begin{figure}[htp]
    \centering
    \includegraphics[width=0.45\textwidth]{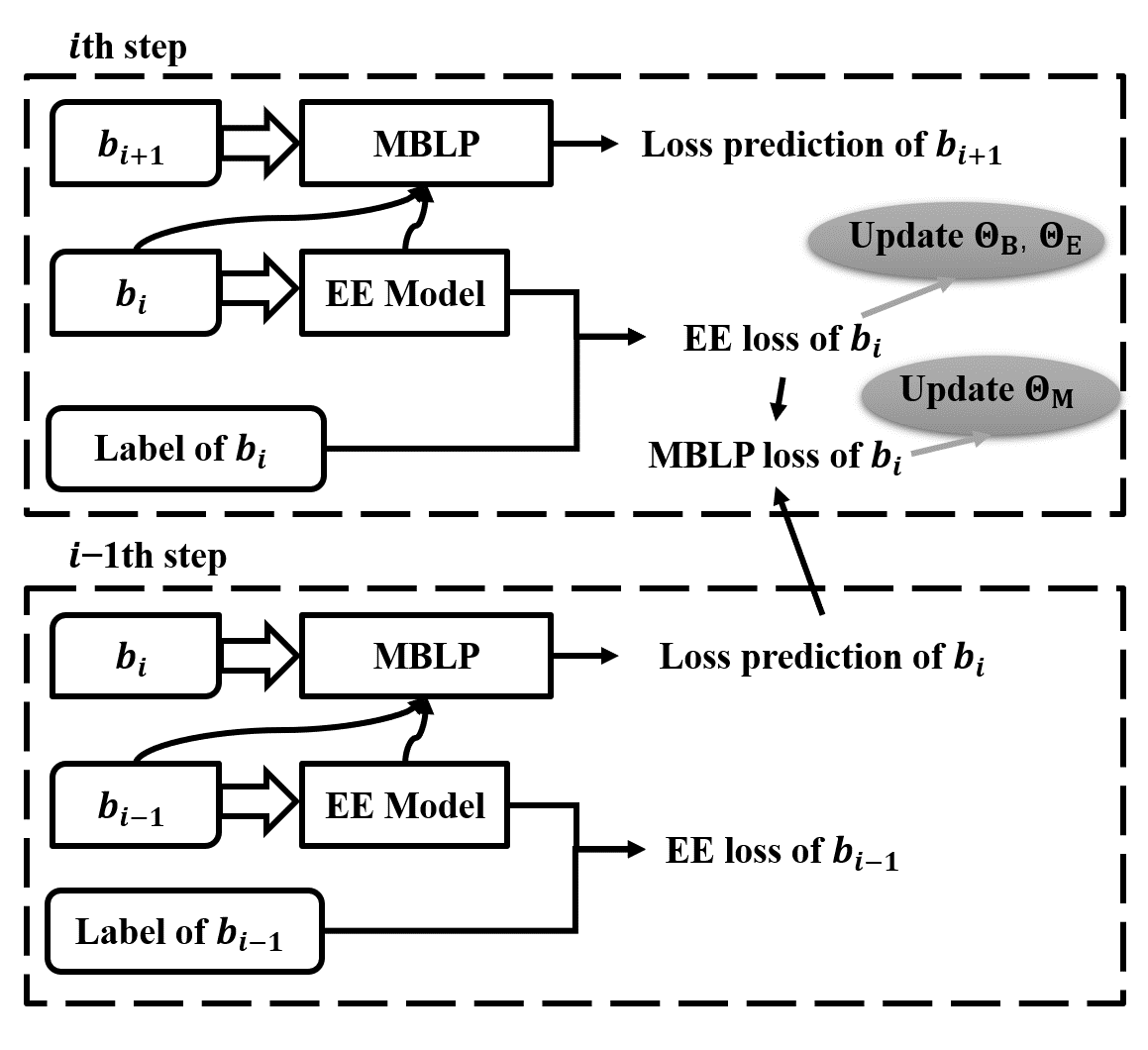}
    \caption{Delayed training strategy. }
    \label{fig:delayed}
\end{figure}

We propose a delayed training strategy for our method, and the sketch map is shown in Figure \ref{fig:delayed}.

For MBLP, we hope that the loss prediction of an unlabeled samples is based on the selected samples from previous batches.
The target loss of an unlabeled sample is the loss after the previously selected sample has participated in the training.

We divide the labeled data into fixed-size batch for training, and the $i$-th batch is recorded as $b_i$.
At the $i$-th step of training, we first initialize SMM with learnable parameters.
Next, we regard $b_{i}$ as the selected sample to update SMM with $b_{i}$, and then predict and store the loss of samples in $b_{i+1}$ for next step.
Then we use the current EE model to predict samples in $b_{i}$, and calculate $\{\mathcal{L}_{ee}^S\}_{S\in b_i}$ as the EE loss. $\{\mathcal{L}_{ee}^S\}_{S\in b_i}$ will be used to update the EE model, and supervise the loss prediction of $b_{i}$ produced by the $i-1$ step.

Specifically, when supervising MBLP, we consider two objective functions, namely, the mean square error (MSE) between predicted losses and real losses, and the ranking losses of predicted losses. 
For a sample $S$, we sort each prediction in $S$ by real balanced loss, and the prediction with the $j$-th largest loss is denoted as ${S^j}$. The MSE between predicted losses and real losses is
\begin{equation}
\label{mse}
\setlength\abovedisplayskip{0.1cm}
\setlength\belowdisplayskip{0.1cm}
    \mathcal{L}^S_{mse} =  \sum\nolimits_{j} (L'(S^j) - L'_{pred}(S^j))^2
\end{equation}
where $L'$ means the real balanced loss, $L'_{pred}$ is the balanced loss prediction.
the internal ranking loss is
\begin{equation}
\setlength\abovedisplayskip{0.1cm}
\setlength\belowdisplayskip{0.1cm}
\label{eq:sort}
    \mathcal{L}^S_{rI} = \sum_{j=1}^{\bm{N}(S)-1} [L'_{pred}(S^{j+1}) - L'_{pred}(S^{j})]_+ 
\end{equation}

where $[x]_+$ means if $x < 0$, $[x]_+=0$, else $[x]_+=x$. 
Then we use the mean value of the largest $m$ losses in $S$ as the loss of $S$, and calculate the external ranking loss $\mathcal{L}^{b_{i}}_{rE}$ of batch $b_{{i}}$ with reference to Eq.\ref{eq:sort}.

Finally, we use stochastic gradient descent (SGD) to update the learnable parameters, where $\{\mathcal{L}_{ee}^S\}_{S\in b_i}$ is used to update the  parameters in EE and fine tuning sample encoder, $\mathcal{L}^{b_{i}}_{rE} + \sum_{S\in b_{i}} (\mathcal{L}^S_{rI} +\mathcal{L}^S_{mse})$ is used to update parameters in MBLP. 
The whole process is shown in  Algorithm \ref{algo1}.

\begin{algorithm}[t]
\caption{\textbf{Delayed Training Process}} 
\label{algo1}
 {\bf Initialization:} 
\\
    Initialized parameters: sample encoder $\mathbf{\Theta}_B$, event extraction model $\mathbf{\Theta}_E$, MBLP model $\mathbf{\Theta}_M$;\\
    Initialized dataset: labeled sample set $\mathcal{L} = \emptyset$, unlabeled sample set $\mathcal{U}$. Randomly select $U_0$ from $\mathcal{U}$, $\mathcal{U}=\mathcal{U} \backslash U_t$, $L_t = label(U_t)$, $\mathcal{L} = \mathcal{L} \cup L_t$.
\begin{algorithmic}[1] 
\State some description 
\For{$t=1$ to query times} 
    \State Split $\mathcal{L}$ into fixed size batch set $B = \{b_1,b_2,b_3,...\}$.
   
    \For{$i<$ number of batches in $B$}
        \State Calculate $\{\mathcal{L}_{ee}^S\}_{S\in b_i}$.
        \If{$i > 1$} 
            \State Calculate $\mathcal{L}^{b_{i}}_{rE} + \sum_{S\in b_{i}} (\mathcal{L}^S_{rI} +\mathcal{L}^S_{mse})$ according to (\ref{mse}) and (\ref{eq:sort}), and minimize it to update $\mathbf{\Theta}_M$.
        \Else 
            \State Pass. 
        \EndIf 
        \State Initialize SMM, update SMM by samples in $b_{i}$ according to (\ref{SMM}).
        \State Predict and store the losses of samples in $b_{i+1}$.
        \State Minimize $\{\mathcal{L}_{ee}^S\}_{S\in b_i}$ to update $\mathbf{\Theta}_B$ and $\mathbf{\Theta}_E$.
        
    \EndFor 
    \State \textbf{endfor}
    \State Split $\mathcal{U}$ into fixed size batches $B = \{b_1,b_2,b_3,...\}$.
    \State Initialize SMM, $U_t=\emptyset$.
    \For{$i<$ target query number}
        \State Select a sample $S$ from $b_i$ by inter-exter loss ranking method and add it into $U_t$.
        \State Update SMM by $S$ according to (\ref{SMM}).
    \EndFor 
    \State\textbf{endfor}
    \State $\mathcal{U}=\mathcal{U} \backslash U_t$, $L_t = label(U_t)$, $\mathcal{L} = \mathcal{L} \cup L_t$, shuffle $\mathcal{L}$, shuffle $\mathcal{U}$.
 \EndFor 
 \State \textbf{endfor}
\end{algorithmic} 
\end{algorithm}


%% file: experiment.tex
\section{Experiments}
\begin{table}[h]
    \centering
    \resizebox{0.43\textwidth}{!}{%
    \begin{tabular}{@{}lccc@{}}
    \toprule
    datasets                   & size  & event types & argument roles \\ \midrule
    military EE  & 12064 & 13 &  24      \\ 
    legal EE     & 2240  & 11  &  30    \\ 
    financial EE & 5082  & 10  &  36      \\ \bottomrule
    \end{tabular}%
    }
    \caption{Information statistics of three datasets.}
    \label{tab:datasets}
\end{table}
To verify the effectiveness of our method and its practical value in domain EE, we perform the experiments on three Chinese domain EE datasets. 
As the basic unit of Chinese is Chinese characters, it is a challenging task to extract Chinese events.
Considering this, using Chinese event extraction data sets can fully demonstrate the effect of our model. 
The details of these datasets are summarized in Table \ref{tab:datasets}.
These three datasets are labeled by domain experts according to the requirements, and we publish our code and datasets on Github\footnote{For the reason of double-blind, we first submit the code and data to CMT system.}.
Our experiments mainly verify the following points:
(1) The effectiveness and stability of our active learning based event extraction method.
(2) The superiority of the internal-external sample loss ranking method on sample importance evaluation.
(3) The superiority of our MBLP and batch-based sample selection strategy.


\subsection{Experimental Setting}

We compare our method other active learning based event extraction method. And we use the random selection as a baseline.
The models participating in the comparison experiment are as follows:
\begin{itemize}
    \item \textbf{Random} is the baseline which random select sample from $\mathcal{U}$ in each round.
    \item \textbf{Uncertainty} \cite{wonsild2020danish} is the method which select samples based on the uncertainty of prediction.
    \item  \textbf{Diversity} \cite{wonsild2020danish} is the method which select samples based on sample diversity.
    \item \textbf{Uncert+Diver}\cite{zhuang2020active} refers to the method of sample selection consider prediction uncertainty and sample diversity at the same time.
    \item \textbf{Loss\_Pred} \cite{yoo2019learning} is the method that uses an end-to-end loss prediction model to evaluate sample importance. 
    \item \textbf{Our} means the method proposed in this paper with MBLP and the batch-based sample selection strategy based on internal-external loss ranking.
\end{itemize}


For each dataset, we randomly select 70\% of the samples as unlabeled pool $\mathcal{U}$ for sample selection, and use the rest as the testing data. 
We select 100 samples each time according to our selection strategy and add them to the training set to train the EE model.
In our experiments, the termination condition of active learning process is that all samples in $\mathcal{U}$ are added to $\mathcal{L}$.
In our method, $m$ is 10 and the size of $d_M$ and other hidden vector in our model is 256.
We use open source sample encoder initialization parameters \footnote{https://huggingface.co/hfl/chinese-bert-wwm-ext}.
We show the performance of each method by drawing the learning curve of the F1 value of the EE model on the independent test set.
To report the mean and standard deviation of performance,
the experiment repeats for five times.

\begin{table}[htp]
    \centering
    \resizebox{0.45\textwidth}{!}{%
    \begin{tabular}{@{}lccc@{}}
    \toprule
    
    Datasets              &legal     &financial     & military  \\ \midrule
    Random              &4.32/5.19 &5.31/5.60     &4.58/5.27     \\ 
    Uncertain           &2.41/3.23 &3.15/3.74     &3.41/4.28     \\ 
    Diversity           &1.97/2.83 &3.26/4.03     &3.25/3.79    \\ 
    Uncert+Diver        &1.78/2.06 &3.22/3.62     &2.70/3.50     \\ 
    Loss\_Pred          &1.86/\textbf{1.73} &2.44/2.57     &3.04/3.27    \\ \midrule
    Our                 &\textbf{1.61}/1.88 &\textbf{1.94}/\textbf{1.83} &\textbf{2.37}/\textbf{2.93}  \\ 
   \bottomrule
    \end{tabular}%
    }
    \caption{The standard deviation of F1 score (trigger/argument) with 20\% samples are labeled.}
    \label{tab:std}
\end{table}

\begin{table}[hbp]
    \centering
    \resizebox{0.45\textwidth}{!}{%
    \begin{tabular}{@{}lccccc@{}}
    \toprule
    Dataset &\multicolumn{5}{c}{legal} \\
    \cmidrule{2-6}
    Percentage               & 10\%  & 20\% & 30\% & 40\% & 50\% \\ \midrule
    Loss\_Pred+mean   &62/71 &75/75 &78/80 &80/84 &81/87      \\ 
    Loss\_Pred+IE       &63/72 &77/78 &80/83 &81/85 &82/88    \\ 
    MDLP+batch          &66/71 &77/79 &79/84 &81/86 &82/89     \\ \midrule
    Our                 &\textbf{69}/\textbf{75} &\textbf{79}/\textbf{80} &\textbf{82}/\textbf{85} &\textbf{83}/\textbf{87} &\textbf{83}/\textbf{90} \\ 
    \midrule
    
    Dataset &\multicolumn{5}{c}{financial} \\
    \cmidrule{2-6}
    Percentage               & 10\%  & 20\% & 30\% & 40\% & 50\% \\ \midrule
    Loss\_Pred+mean   &69/68 &71/71 &75/72 &76/74 &76/78      \\ 
    Loss\_Pred+IE       &70/68 &72/73 &76/73 &77/75 &78/78    \\ 
    MDLP+batch          &69/70 &74/73 &76/74 &78/76 &78/79     \\ \midrule
    Our                 &\textbf{71}/\textbf{70} &\textbf{75}/\textbf{74} &\textbf{76}/\textbf{77} &\textbf{78}/\textbf{78} &\textbf{78}/\textbf{79}\\
    \midrule
    Dataset &\multicolumn{5}{c}{military} \\
    \cmidrule{2-6}
    Percentage               & 10\%  & 20\% & 30\% & 40\% & 50\% \\ \midrule
    Loss\_Pred+mean   &55/56 &62/63 &68/67 &70/71 &71/72      \\ 
    Loss\_Pred+IE       &60/59 &70/71 &74/75 &75/77 &78/79    \\ 
    MDLP+batch          &59/58 &71/70 &75/76 &78/79 &79/80     \\ \midrule
    Our                 &\textbf{63}/\textbf{65} &\textbf{79}/\textbf{78} &\textbf{79}/\textbf{81} &\textbf{80}/\textbf{81} &\textbf{80}/\textbf{81}\\ 
    \bottomrule
    \end{tabular}%
    }
    \caption{The F1 score (trigger/argument) with percentage data.}
    \label{tab:ablat}
\end{table}

\begin{figure*}[ht]
\centering

\subfigure[Results on legal dataset.]{
\begin{minipage}[t]{0.32\linewidth}
\centering
\includegraphics[width=0.9\linewidth]{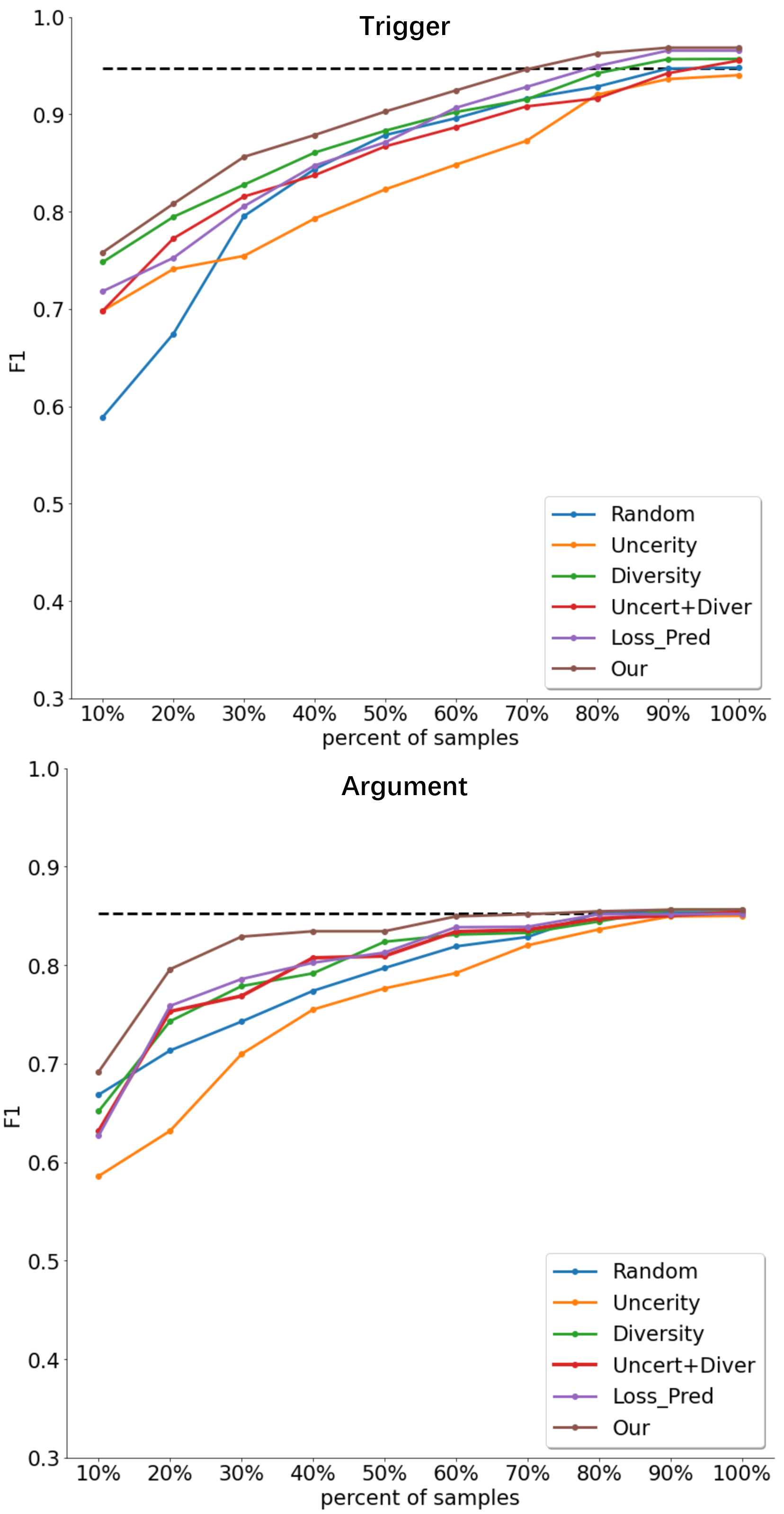}
\end{minipage}%
}%
\subfigure[Results on financial dataset.]{
\begin{minipage}[t]{0.32\linewidth}
\centering
\includegraphics[width=0.9\linewidth]{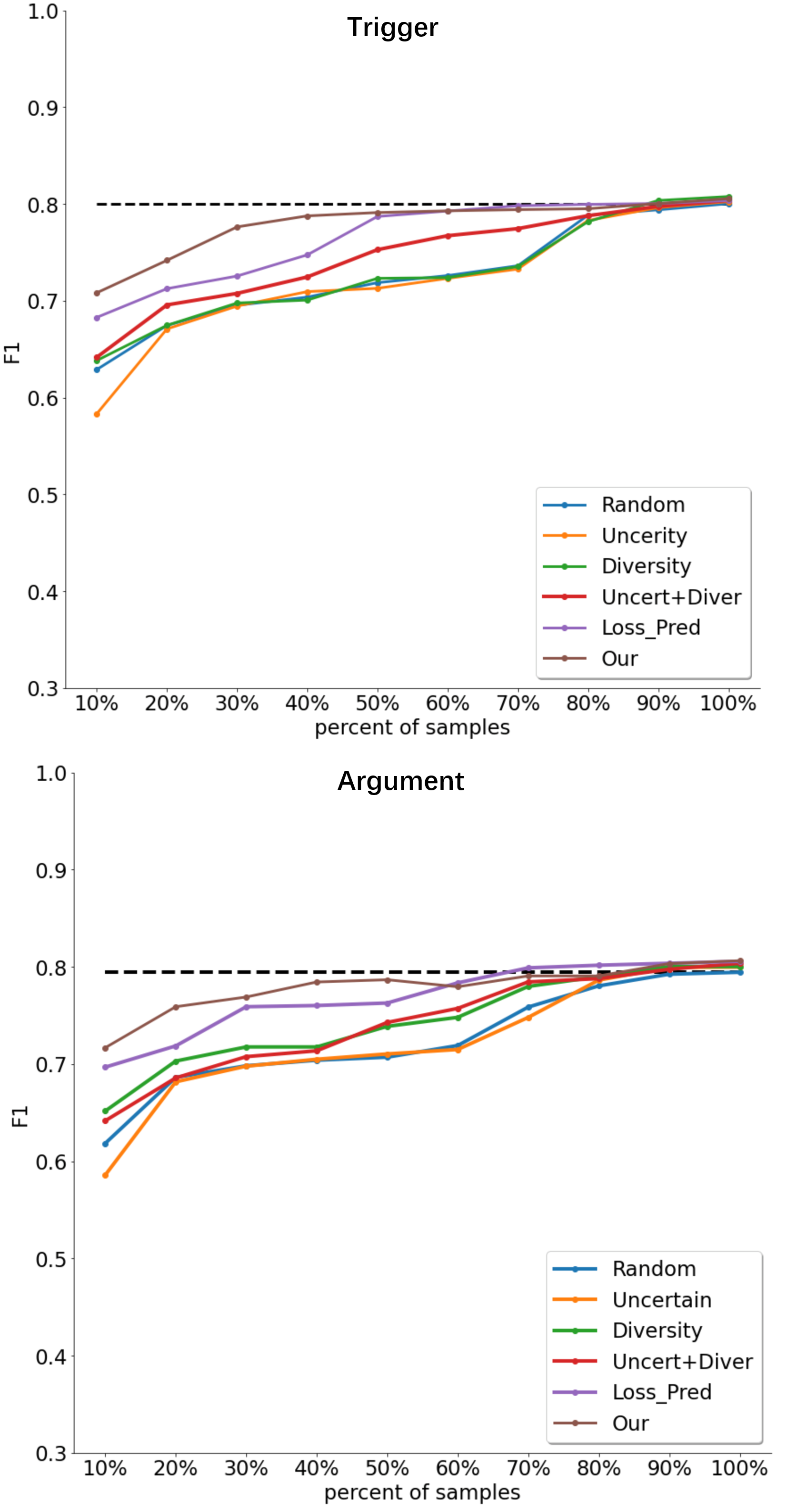}
\end{minipage}%
}%
\subfigure[Results on military dataset.]{
\begin{minipage}[t]{0.32\linewidth}
\centering
\includegraphics[width=0.9\linewidth]{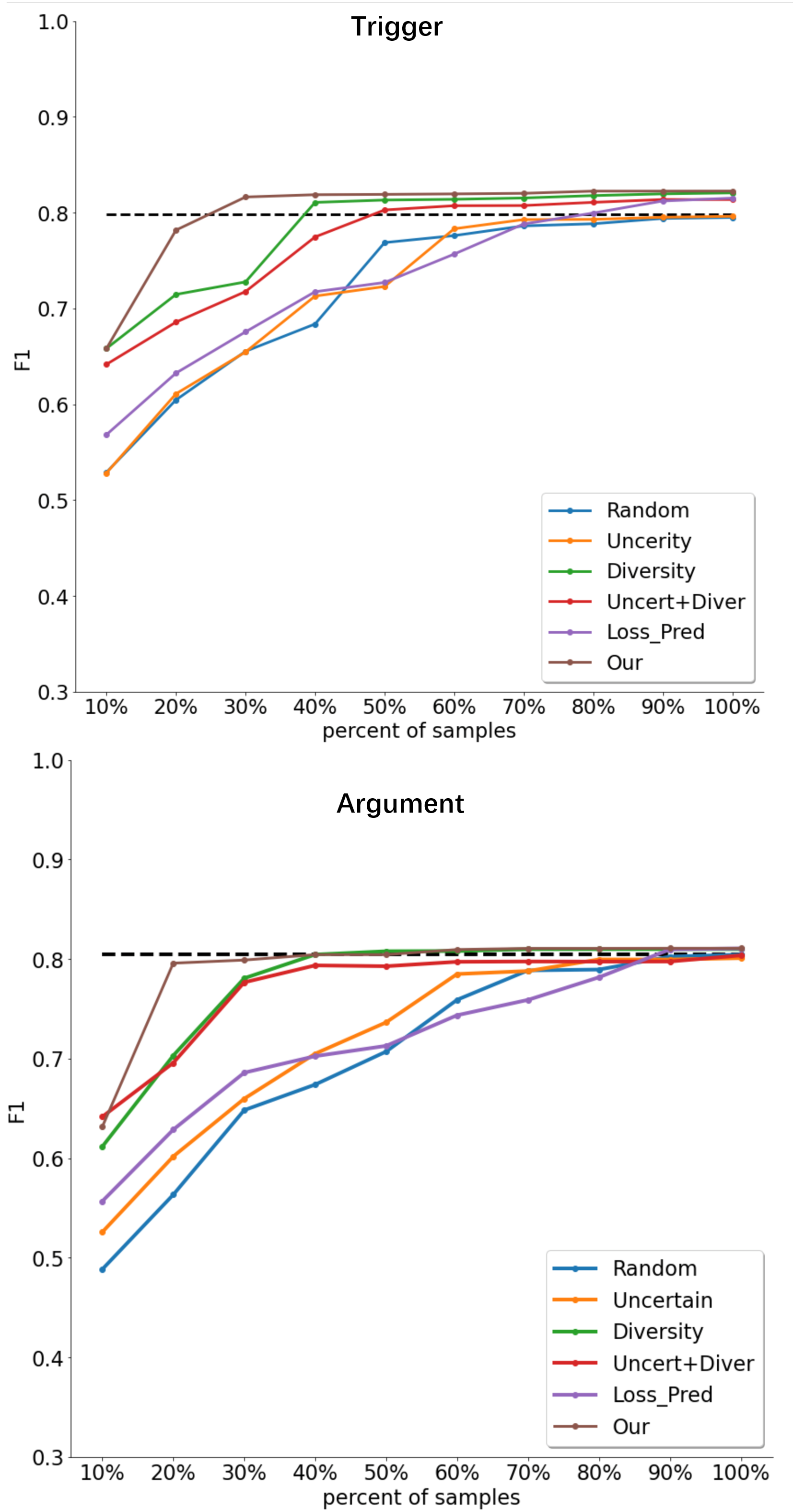}
\end{minipage}%
}%

\caption{Comparisons of different active learning methods in terms of the mean F1 score on three domain event extraction datasets. The dashed line represents the F1 value when using full data for training. }
\label{allresults}
\end{figure*}

\subsection{Experimental Result}
\subsubsection{General Performance}
Figure \ref{allresults} shows the mean F1 scores of using different active learning methods on three datasets.
It can be observed that when there are few training samples, the event extraction model performs poorly, which is due to the lack of task-specific features in the labeled samples, which leads to the poor performance of the model on the test set.
However, our method can select samples which are more valuable for EE model, rapidly improving the quality of training set.
So, our method outperforms other methods with the selecting the same number of samples on both trigger classification and argument classification under all datasets. 
It prove the effectiveness of the combination of MBLP and our new sample selection strategy on event extraction task.

In addition, compared with other methods, the performance of EE with our method is improved faster.
This shows that the samples selected by our method are (1) more effective in improving the performance of the EE model, and (2) more informative.
On financial data and military data, the heuristic method does not significantly improve the performance of the model when the number of samples is small.
This shows that heuristic methods are difficult to adapt to complex deep neural networks.
Compared with the \textbf{Loss\_Pred}, adding the memory loss prediction method can effectively improve the richness of information in the selected sample, and makes the performance of the EE model improve faster.
We also recorded the standard deviation of F1 value in many experiments, and the results show that our model is also superior to other methods in stability.
Table \ref{tab:std} shows the standard deviation of F1 score with 20\% samples are labeled for training. Our method has always maintained a low standard deviation.
It can be seen that the heuristic method is unstable in EE task, which is because the existing sample selection methods based on diversity and uncertainty ignore the local information of samples, resulting in obvious fluctuation of sample quality.
\textbf{Loss\_Pred} is instability on military dataset. 
The reason is that military dataset has more event types, greedy selection method can not distribute the sample distribution of different events well. 

\subsubsection{Ablation Study}


In this section, we study the effectiveness of each part of our method.
We compare our whole method with following settings:
\begin{itemize}
    \item \textbf{Loss\_Pred+mean}: In this method, we use a conventional loss prediction method \cite{yoo2019learning}, and use the average of all losses in the sample as its final loss. In the selection process, we use greedy selection strategy.
    \item \textbf{Loss\_Pred+IE}: In this method, we use a conventional loss prediction method \cite{yoo2019learning}, and use our internal-external loss ranking method to evaluate the importance of sample.In the selection process, we use greedy selection strategy.
    \item \textbf{MBLP+batch}: This is the method of removing internal-external loss ranking from our whole method.
\end{itemize}

Table \ref{tab:ablat} shows the results of ablation study.
It can be observed that the methods with MDLP and batch-based sample selection strategy improve rapidly in early stage, especially in the legal and military domains.
It indicates that the sample set selected base on MBLP contains richer and more valuable information, and when there are more types of events, the advantages are more obvious.

The methods with inter-exter loss ranking perform better.
In particular, the military domain contains more argument roles, and the information in a sample is more complicated, the methods with inter-exter loss ranking can effectively select more important samples for training. 
This is because our methods considers more abundant local semantic information and filters out unimportant noise through internal loss ranking and selection.
As conclusion, the ablation study prove the superiority of each module in our methods.


\subsubsection{Parameter Study}
In our method, $m$ is an important hyperparameter.
The value of $m$ may affect the performance of the model. 
When $m$ is small,some information in the sample may be ignored, and when $m$ is large, unimportant information may interfere with selection.
Therefore, we examined our method under different $m$ conditions.

\begin{table}[h]
    \centering
    \resizebox{0.45\textwidth}{!}{%
    \begin{tabular}{@{}lcccccccc@{}} 
    \toprule
    
     $m$               & =6  & =7 & =8 & =9 & =10  & =11 & =12& $=+\infty$ \\ \midrule
    Legal(\%)              &37 &30 &26 &23 &\textbf{21} &22 &25  &31  \\ 
    Financial(\%)      &25 &18 &14 &12 &\textbf{11} &13 &16  &22 \\ 
    Military(\%)        &41 &33 &29 &22 &19 &\textbf{17} &18  &30 \\ 
    \bottomrule
   
    \end{tabular}%
    }
    \caption{The percentage of data used by the model to reach 90\% of the average F1 under the training with full data. $+\infty$ means we will consider all predictions in a sample.}
    \label{tab:m}
\end{table}
As show in Table \ref{tab:m}, when $m$ increases, the speed of reaching the same training goal first becomes faster and then slower, indicating that our previous analysis is correct. 
In the legal and financial domains, when $m=10$, the EE model improves the fastest. In the military domain, the effect is best when $m=11$, which may be caused by more event types in this domain, and more complex information contained in the military samples.

%% file: conclusion.tex
\section{Conclusion}
This paper analyzed the shortcomings of the previous AL methods, and proposed a new AL framework for EE.
We proposed a MBLP model and a batch-based selection strategy.
MBLP predicts the loss of the unlabeled sample by the current state of EE model and the information of selected samples.
We proposed an inter-exter loss ranking method to evaluate the importance of samples according to important local information.
Our method improve the overall quality of sample selection and reduce the computational cost in selection process.
We also proposed a delayed training strategy to train the EE model and MBLP together. 
Extensive experiments on EE datasets in three domains verified the effectiveness and reliability of our method.

In the future, we will study how to make our method more robust to the risks of AL sample selection.
For example, existing AL methods ignore the potential errors in the prediction results with high confidence . 
In this case, the incorrect knowledge in the model may not be corrected, thus affecting the training process of AL.
